\documentclass[fleqn,10pt]{scrartcl}
\usepackage[utf8]{inputenc}
\usepackage{amsmath} 
\usepackage[a4paper,inner=2.5cm,outer=2cm,bottom=2.5cm,top=2.5cm]{geometry}
\usepackage{amsfonts} 
\usepackage{amssymb} 
\usepackage[font=sf]{caption}
\usepackage[T1]{fontenc} 
\usepackage{nameref} 
\usepackage{booktabs} 
\usepackage{blindtext}
\usepackage{array}
\usepackage{longtable}
\usepackage{lscape}
\usepackage{here}
\usepackage{caption}
\usepackage{subcaption}
\usepackage{tabularx}
\usepackage{printlen}
\newcolumntype{R}[1]{>{\raggedleft\arraybackslash}p{#1}}
\newcolumntype{C}[1]{>{\centering\arraybackslash}p{#1}}
\newcolumntype{L}[1]{>{\raggedright\arraybackslash}p{#1}}
\usepackage{setspace}

\usepackage{color}
\usepackage{wrapfig}
\usepackage{rotating} 
\usepackage{lscape}
\usepackage{graphicx}
\usepackage[numbers]{natbib}
\usepackage{url}

\usepackage{pdfpages}
\usepackage[table,xcdraw]{xcolor}
\usepackage{svg}
\usepackage{multirow,tabularx}
\usepackage{colortbl,color}
\definecolor{Gray}{gray}{0.9}
\usepackage{tabto} 
\usepackage{tikz}
\usepackage{adjustbox}
\usepackage[miktex]{gnuplottex}    
\usepackage{pgfplots}
\pgfplotsset{compat=1.7}

\title{From Motion to Muscle}

\begin{document}
\date{}
\setlength{\parindent}{0pt}
\setlength{\parskip}{1em}
\thispagestyle{empty}

\large\textbf{\text{Marie D. Schmidt*\textsuperscript{1}\textsuperscript{,}\textsuperscript{2}, Tobias Glasmachers\textsuperscript{2}, Ioannis Iossifidis\textsuperscript{1} }} \\\\ 
\small\text{\textsuperscript{1}Institute of Computer Science, University of Applied Science Ruhr West, Mülheim an der Ruhr, Germany}\\
\small\text{\textsuperscript{2}Institute for Neural Computation, Ruhr University, Bochum, Germany}\\
\small\text{marie.schmidt@hs-ruhrwest.de}\\
\small\text{tobias.glasmachers@ini.rub.de}\\
\small\text{ioannis.iossifidis@hs-ruhrwest.de}

\section*{Abstract}
Voluntary human motion is the product of muscle activity that results from upstream motion planning of the motor cortical areas. We show that muscle activity can be artificially generated based on motion features such as position, velocity, and acceleration. For this purpose, we specifically develop an approach based on a recurrent neural network trained in a supervised learning session; additional neural network architectures are considered and evaluated. The performance is evaluated by a new score called the zero-line score. The latter adaptively rescales the loss function of the generated signal for all channels by comparing the overall range of muscle activity and thus dynamically evaluates similarities between both signals. The model achieves a remarkable precision for previously trained motion while new motions that were not trained before still have high accuracy. Further, these models are trained on multiple subjects and thus are able to generalize across individuals. In addition, we distinguish between a general model that has been trained on several subjects, a subject-specific model, and a specific pre-trained model that uses the general model as a basis and is adapted to a specific subject afterward. The subject-specific generation of muscle activity can be further exploited to improve the rehabilitation of neuromuscular diseases with myoelectric prostheses and functional electric stimulation.

\raggedbottom 
\pagenumbering{arabic}  
\newpage
\setcounter{page}{1}    
\section{Introduction}\label{sec:Introduction}  
Every day, we use our arms and hands to move precisely while flexibly interacting with our environment. Human motion execution is the product of muscle contraction caused by muscle activity which, in turn, results from upstream motion planning of the motor cortical areas. Motion is defined as a change in position over time, and this can be described by parameters such as time, direction, and velocity. In the 1980s, Georgopoulos and colleagues found a correlation between the movement direction of the hand and the motor cortical activity \citep{georgopoulos1982relations, georgopoulos1983spatial}. Moreover, speed and, with a less prominent effect, acceleration and position are continuously represented in motor cortical activity during reaching \citep{ashe1994movement, moran1999motor}. There is some controversy about whether the motor cortex represents so-called high-level features of the hand as described above (direction, speed, and acceleration) or low-level features for muscle groups such as muscle activity and force motivated by \citep{todorov2000direct, mussa1988neurons, evarts1968relation}. Churchland and colleagues developed a dynamical system approach to better understand the neural activity in the motor cortex \cite{churchland2012neural}. Further, the motor cortex might be explained by utilizing a recurrent neural network (RNN) \citep{michaels2016neural, sussillo2015neural, stroud2018motor} which in itself exhibits dynamical behavior. These models show that preparatory activity sets initial conditions that unfold predictably to control muscles during reaching. We might assume that the preparatory activity draws on pre-learned inverse dynamics that generate the associated muscle activity with measurable angular position, velocity, and acceleration for each joint.

Building on this thesis, we demonstrate that muscle activity can be generated artificially for known and unknown motion based on high-level motion features for each joint (or for the hand instead), which is similarly represented in our brain. For this, we develop a recurrent neural network with long-short term dependencies in a supervised learning session with motion parameters such as angular position, velocity, and acceleration of the arm. Previously trained motion can be generated with a remarkable precision, while new motions that are not previously trained reach a high precision in some cases. It has to be clarified that we do not aim to represent specific cortical activity but to show that muscle activity measured directly on the surface of the arm can be predicted based on motion parameters.

This concept is based on the assumption that motion parameters such as angular motion, velocity, and acceleration, have a context to muscle activity. The specifically developed recurrent network is the most promising model for generating time series data due to its ability to recognize time dependencies in the motion sequence and thus is the canonical candidate to approach the relation between motion and muscle activity. More precisely, a Long Short-Term Memory (LSTM) network, which is a type of recurrent neural network, allows for stimulation from earlier input remaining as hidden states to influence predictions at the current time step. This enables recurrent networks to exploit a dynamically contextual window which then can be utilized for time dependency of muscle activity. In addition to the recurrent network, we consider and evaluate other neural network architectures. A vanilla feedforward neural network (FNN) is a base model which allows the signal to only travel in one direction from input to output. Further, a convolutional neural network (CNN) is introduced, which extracts spatial and temporal dependencies of the time series by applying different sized kernels and filters using the whole movement (and not one point in time) as input. Thereby, the motion data as a whole is mapped to the muscle activity. The performance of a model is presented by the mean square error (MSE) loss function of the neural network. We further propose a new better-suited measure, the zero-line score (Z-score), that adaptively rescales the loss function of the muscle activity and compares the generated signal to the overall range of the muscle activity and thus identifies similarities between both signals. 

Mandatory for all kinds of applications is the generalization properties of the architecture, along with the question of how transferable the feature is. We aim for a model which can further extrapolate across multiple subjects as well as for new motions. To provide a measure for generalization, we evaluate different disparate motions and transfer learning across all feature combinations across multiple subjects. The muscle activity, recorded by electromyography (EMG), is known as a high inter-subject variable due to varying physiological factors, slightly different electrode placement, and other skin conditions \citep{sheng2019common, farina2014extraction, araujo2000inter, nordander2003influence, hogrel1998variability, farina2002influence}. Therefore, we evaluate models that have been trained on different individuals: We distinguish between a general model trained on multiple subjects but exclude the data from the subject that is referred to test these models, a subject-specific model that is trained entirely on this one subject, and a specific, pre-trained model that uses the general model as a basis and is then adapted to the specific subject. 

To evaluate the true generativity of our approach, we go beyond generating muscle activity based on already known motions and predict new, unseen motions covering a huge range of motions. Last, we fall back on the input parameters we have chosen: angular position, velocity, and acceleration. We further distinguish whether all these motion parameters are crucial input data for generating artificial muscle activity or whether the redundancy of velocity and acceleration due to their reproducibility by deriving the position can be observed as well. In addition, we check if the position and orientation for the hand alone, also known as end-effector (EEF), are sufficient to drive the muscle activity of the upstream arm joints/segments, which can be described by inverse kinematics based on the EEF.

The subject-specific adjustment is particularly advantageous for myoelectrical controlled systems such as prostheses or exoskeletons. Most control systems rely on the classification of residual signals. Especially for training deep learning models, large data sets are needed, otherwise the performance and quality of generalization will degrade due to lack of data, hindering real-world applications. The concept of our muscle activity generation can be adapted to provide these applications with additional subject-specific data and increase their performance. Furthermore, in a rehabilitation context, movements can be supported by targeted functional electric stimulation (FES) based on the generated muscle activity.

\section{Material and Methods}\label{sec:Material/Methodes}  
In this section, the entire process from data acquisition to pre-and post-processing to the construction of the different neural network architectures and their hyperparameter tuning is outlined. 

\subsection{Experimental protocol} 
In total, five healthy subjects ($2$ female, $3$ male in the age of $26~ \pm 2$ years) participated in the experiment. They all have given their written consent to the study. The study involving human subjects was reviewed and approved by the Ethics Committee of the Ruhr-University Bochum. Each subject performed $20$ tasks with $18$ repetitions of each isotonic movement resulting in $360$ motion sequences per subject. An isotonic movement is caused by a muscular contraction that leads to a change in muscle length and thereby causes a motion at the corresponding joint. The movements can be categorized into three groups; simple motion, combined, and complex motion. The simple motions include shoulder flexion, shoulder extension, shoulder abduction, elbow flexion, elbow flexion with a supinated forearm, wrist flexion, wrist extension, and wrist pronation. The combined movements are composed of shoulder abduction elbow flexion, shoulder flexion elbow flexion, and shoulder abduction wrist extension. The complex movements try to mimic everyday activity's as are breaststroke, relay handover, reading a clock, diagonal reach, waving gestures, and pointing into three points in space. 

\begin{figure}[H]
\centering
\includegraphics[width=\linewidth]{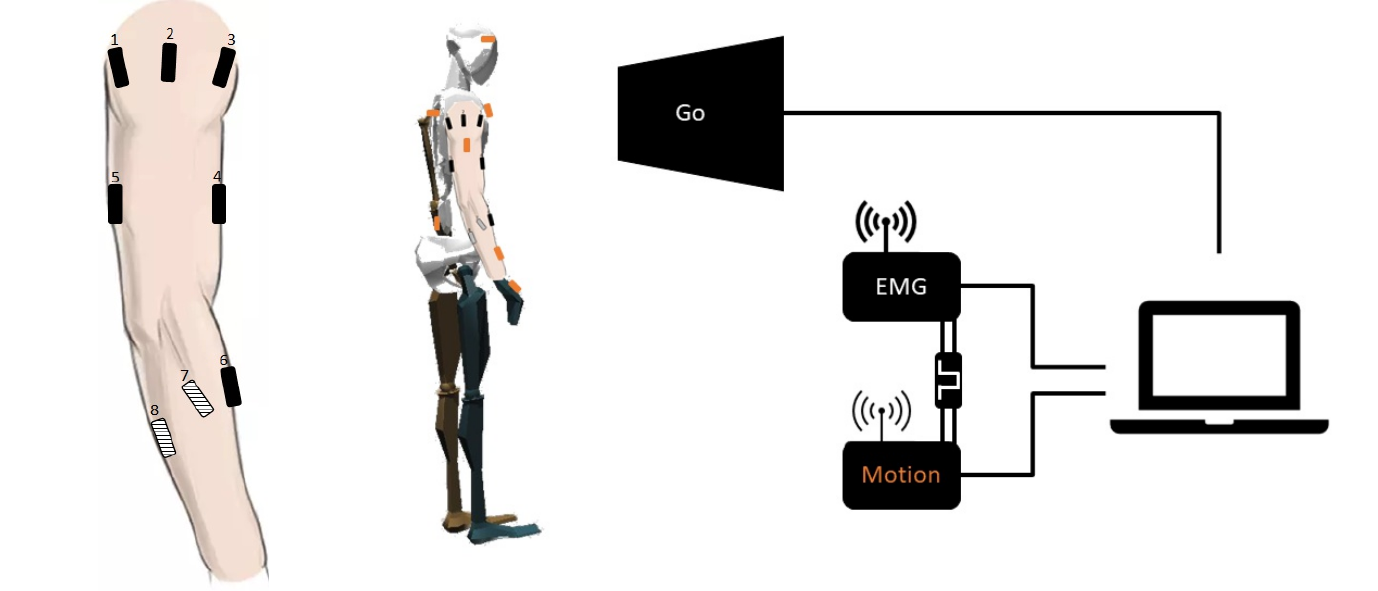}
\put(-470,1){\textsf{a)}}
\put(-350,1){\textsf{b)}}
\caption{(a) Overview of the electrode (rectangles) placement on the right upper limb (deltoid posterior (1), lateral (2) and anterior(3), biceps short head (4), triceps brachii lateral head (5), extensor carpi radialis longus (6), pronator teres (7), and flexor carpi ulnaris (8) with the last two shaded), modified from \citep{arm2021arm}. (b) Setup up including Delsys Trigno EMG System and Xsens motion capture system (motion sensor in orange) connected to the the trigger module and the laptop. }
  \label{fig:electrodePlacement}
\end{figure}

The muscle activity of these movements is recorded at $2222$ Hz using the Trigno Wireless EMG System (Delsys Inc., Boston, MA, USA) with two Quadro electrodes. The skin preparation and placement of electrodes were performed according to the recommendation of the SENIAM manuscript \citep{konrad2005emg}. The EMG electrodes were placed on the upper right arm on the following muscles: deltoid anterior, medial, and posterior, biceps short head, triceps brachii lateral head, pronator teres, flexor carpi radialis, and extensor carpi ulnaris (Fig.:\ref{fig:electrodePlacement}). The EMG system is synchronized with the motion tracking from Xsens Motion Capture via the Delsys trigger box. The Xsens Motion Capture system (Xsens Technologies B.V., P.O. Box 559, 7500 AN Enschede, Netherlands) uses the upper body configuration including $11$ sensors covering both arms and the torso (wrist, forearm, upper arm, shoulder for each side, sternum, pelvis, and head). The application and advanced N-pose calibration of the sensors is performed according to their manual \citep{myn2015xsens}. The Xsens Motion Capture samples with a rate of $60$ Hz.

During the experiment, the subjects stand in front of a screen showing the visual interface which provides the instructions always starting with the resting position for $4$ s (as long as the ‘resting window’ is open). After that, an instruction window pops and refreshes every $7.5$ s indicating the next repetition. During this time the subject is asked to perform the described movement repeatedly. To avoid muscle fatigue $60$ s of rest is granted after each task.

\subsection{Data preprocessing} 
The EMG signal is baseline corrected, and outliers are determined by considering values with six standard deviation of the mean. The detected outliers are mostly related to cable movement and are fitted and subtracted by a spline. Afterwards, the EMG data is smoothed with a root mean square \eqref{eq:RMS} over a window size of $200$ ms and simultaneously downsampled to the Xsens frequency of $60$ Hz. Further, for each subject individually, all EMG and motion data channels are normalized between $0$ and $1$ and $-1$ and $1$, respectively. The normalization is necessary due to huge differences between the individual channels. The Xsens system calculates the position and orientation and the velocities of the body segments based on the recorded acceleration. In addition, the angular position of the joint is given. In this study, the angular position of the shoulder, elbow, and wrist reflect shoulder abduction/adduction, shoulder extension/flexion, elbow extension/flexion, elbow rotation, wrist abduction/adduction, and wrist extension/flexion. The angular position is filtered by a third-order Savitzky-Golay filter \cite{savitzky_smoothing_2002-1} and then two times discretely differentiated to compute the angular velocity and acceleration using the forward difference operator $\Delta f\colon n \mapsto f (n+1)-f(n) $. The angular position, angular velocity, and angular acceleration of each joint serve as input data for the following models. As well as the hand position, orientation for the EEF configuration and with their additional velocity, and acceleration for the EEF\textsuperscript{+} configuration. 
\begin{align}
    \phantom{xxxxxxxxxxxxxxxxxxxxxxxxxxxxx}\text{RMS} = \sqrt{\frac{1}{n}\sum_{i=1}^{n}  x_i^2}
    \label{eq:RMS}
\end{align}

\subsection{Neural Network Models}
We investigate the relationship between motion input data and muscle activity that can be learned by a neural network to generate an artificial muscle activity. The models are trained on angular position, velocity, and acceleration for each joint to predict the corresponding muscle activity. In a matter of supervised learning, we compare the predicted muscle activity to the smoothed \eqref{eq:RMS} recorded muscle activity. The recurrent neural network is compared to two other network types: a basic vanilla feedforward network and a more complex convolutional network. Besides the different architectures, the models are also evaluated on different training approaches. These approaches include training on different subjects' data: distinguishing between a general model that has been trained on several subjects, a general model which was fine-tuned on subject-specific data, and a solely subject-specific model. The architectures are evaluated by a general model trained and tested on multiple subjects (Results \ref{sec:Architecture}). For the evaluation of the different training approaches, one subject is excluded from the training data set to, later on, test the model on unseen subject data (Results \ref{sec:Individual training inputs}). Therefore, the training set consists of $15$ repetitions of $n-1$ motions (all motions except one), leaving $2$ repetitions for $n-1$ motion for the training set and one repetition $n-1$ motion for the validation set. The excluded $n-1$ motion is later used to evaluate the performance to generate new motions (Results \ref{sec:Generating new motion}). 

All networks are implemented with the Keras API \citep{chollet2015keras}. The networks use an adaptive learning rate optimization algorithm called Adam \citep{kingma2017adam} to change the learning rate and weights to reduce the loss. The loss is the prediction error of the neural network computed in our case by the mean squared error loss function \eqref{eq:MSE1}. Through backpropagation, the loss is transferred from one layer to another and the weights are modified depending on the losses so that the loss is minimized. The rectified linear activation function (ReLu) \eqref{eq:ReLu} is used in the hidden layers which describe the transformation from input to output from a node. To prevent overfitting, an early stopping with a patience of $5$ epochs and dropout layers are implemented. The dropout randomly sets input units to $0$ with a frequency of the rate parameter at each step during training time.
\begin{align}
    \phantom{xxxxxxxxxxxxxxxxxxxxxxxxxxxxx}\text{MSE} &= \frac{1}{n}\sum_{i=1}^{n} (y_i-x_i)^2,\label{eq:MSE1}\\[1ex]
    f(x) &= \text{max}(x,0)= \begin{cases}
    x   & \text{if}\;\; x >0, \\
    0    & \text{otherwise}\,
\end{cases}\label{eq:ReLu}
\end{align}

\begin{figure}[H]
\centering
\includegraphics[width=\textwidth]{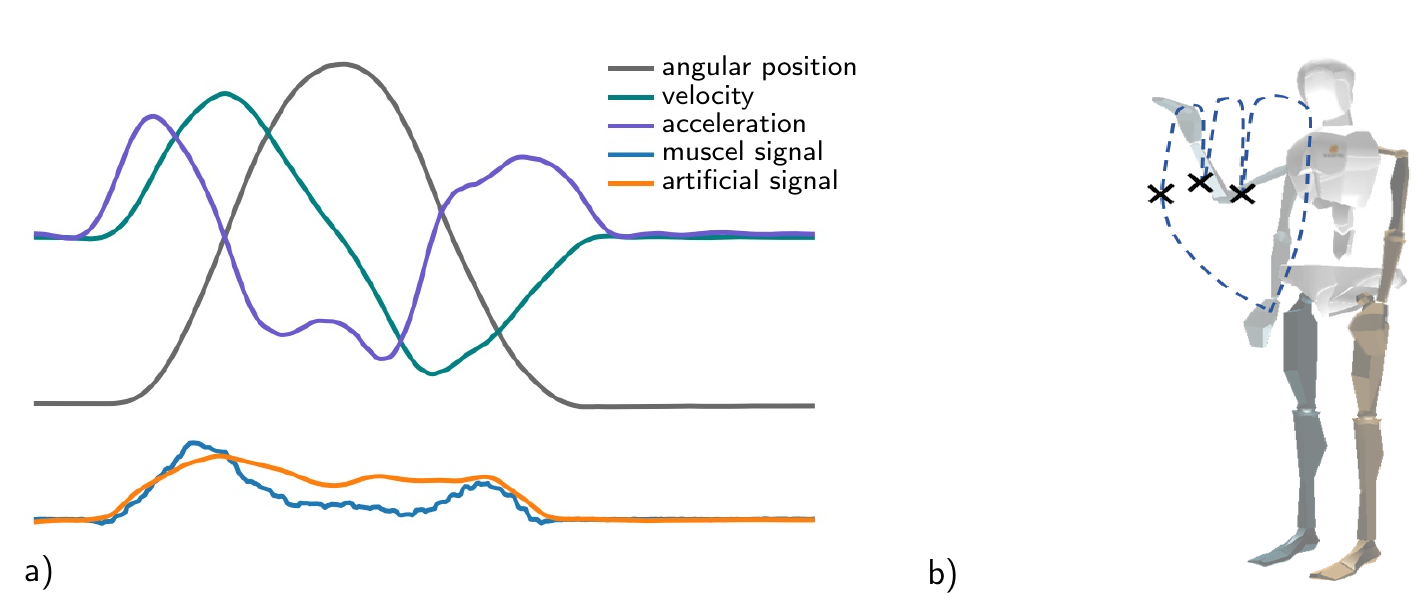}
\caption{(a) Motion trajectory (in terms of angular position, velocity, and acceleration) and muscle biceps signal as well as the artificially generated signal for flexion and extension of the elbow. (b) Visualisation of the new motion, pointing into three points in space (the three points are indicated by the black crosses and trajectory indicated by the dashed line).}
\label{fig:motion3Pmotion}
\end{figure}

\subsubsection{Recurrent Neural Networks} 
In general, recurrent neural networks (RNN) are specialized in time dependencies due to their specific architecture that allows adding information of previous time steps to the current one. Theoretically, all recurrent networks are stateful, however, in Keras, this only applies within a batch. If an additional time sequence is separated into individual batches, the previous information is lost since the states are initialized at the first time step of a batch and reset to 0. Two possibilities are derived from this: either the batch contains the entire sequence of one movement, or the state must be regained during warm-up phase before the actual prediction of the current time step. The latter suggestion can also be used for an online prediction of the muscle activity. In the following, both approaches are applied beginning with the first approach, which trains and predicts the whole sequence at once. 

This recurrent network consists of an input dropout and three hidden Long Short-Term Memory (LSTM) \citep{hochreiter1997long} layers (256, 128, 64 nodes) and a TimeDistrubtion dense output. The LSTM network is a type of RNN that overcomes the vanishing and exploding gradient problem of standard RNNs \citep{hochreiter1997long} by using gates to control the memorizing process. The model is trained with a vector containing (total number of sequences, time steps of each sequence, and number of features). In the LSTM layer, the return sequence and stateful parameter are enabled, which allows the LSTM layer to predict the whole sequence at once. To obtain a good result at the beginning of the sequence, the stateful parameter needs to be enabled. Thus, the memory states and hidden states of the LSTM layer are saved from the former sequence and used as a reasonable starting value for the next sequence. In our context, this works particularly well because every sequence starts and ends at the same resting position (so each sequence has a similar starting value range). However, the downside of this approach is a slower backpropagation due to the length of a whole sequence.

In the second approach, the recurrent network is trained with fractures of the motion sequence, called sub-sequences. Each sub-sequence only consists of a single data point in time, which has to be predicted, and a few prior time steps $i:w+i$ as a warm-up phase, solely used to restore the correct time-dependent states of the recurrent network. Instead of using the full window size of $w$ it is preferable to use just some e.g. $2^i$, which would mean $(1,2,4,8...)$ value of the sub-sequence. In addition, this approach benefits from two models, one for training using the sub-sequence and one for prediction using the whole sequence again allowing different batch sizes for training and testing. While it is more efficient to train with a higher batch size it is necessary to predict with a batch size of one for each time step. Both models (RNN and the sub-sequences model RNNseq) have a similar architecture starting with an input dropout followed by a LSTM layer and a dense output layer. However, in the training model of the LSTM layer, neither the stateful nor return sequence parameter is activated. The return sequence parameter is not needed in this scenario because we are not training on whole sequences. The neglection of the stateful parameter allows a more flexible train regarding the batch size as well as the possibility to shuffle the sub-sequence of the movement. However, to be able to predict a whole sub-sequence in one, a new model needs to be defined using the same weights as for the previously trained model. The architecture only differs in the enabled stateful parameter in the hidden layer and a batch size of one. The states will be rest after a full sequence prediction. This approach can also be used for online generation of muscle activity.

\subsubsection{Feedforward Network} 
The vanilla feedforward network (FNN) is one of the more basic networks with a simple forward pass of information. The architecture is composed of three fully connected hidden dense layers with $512$, $256$ and $128$ nodes, and an $8$-node output according to the number of predicted EMG channels. The model is trained with a batch size of $128$, i.e, the gradient is updated every $128$th sample. In theory, it should be beneficial to have additional information on previous time steps to predict the current time step. Therefore, we develop a vanilla network that is also fed with sequence information (FNNseq) of additional previous time steps $n-i^2=(n-1,n-2,n-4,n-8,..)$, which are added to the feature input vector. This information of previous time steps should improve the prediction of the current step. 

\subsubsection{Convolutional Neural Network} 
Besides the recurrent, the convolutional neural network (CNN) is also able to capture time dependencies through the application of relevant filters.  The CNN works as a feature extractor that transforms the data into a form that is easier to process, with the intention not to lose relevant information  necessary for a good prediction. This is done by the kernel and filter parameters in the convolutional layer. The CNN consists of an input dropout layer with a rate of $0.1$ followed by $5$ one dimensional convolution layers with decreasing number of convolutions $(128, 128, 128, 128, 64)$ and kernel size of $(32, 8, 8, 4, 4)$ and a final output dense layer with $8$ nodes.

\subsection{Hyperparameter Tuning}  
All models were hyperparameter tuned using \texttt{optuna} for \texttt{optkeras} with enabled pruning option on an evolutionary sampler \citep{optuna_2019}. Concerning a minimal validation error the following parameters were optimized: batch size, number of layers, number of nodes, the dropout rate, and number of filter and kernel size for the CNN. 

\section{Results}\label{sec:Results} 
\begin{figure}[H]
\includegraphics[width=\textwidth]{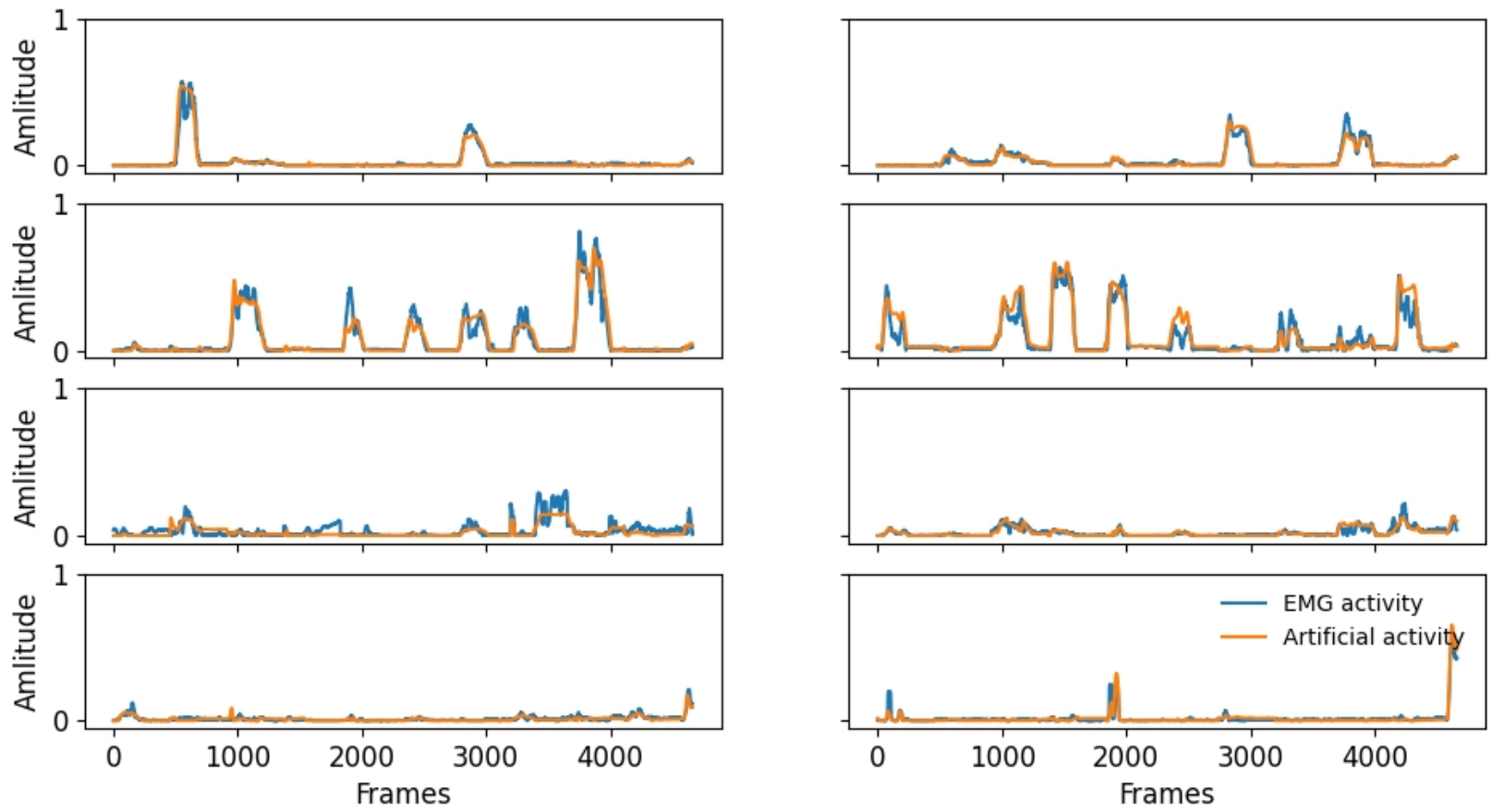}
\caption{Muscle activity for all eight channels (blue) and the artificial generated muscle activity (orange). The artificial activity is generated on the general training approach using a RNN architecture.}
\label{fig:LSTMtestSmall} 
\end{figure}

The artificially generated muscle activity follows the trends of the smoothed original signal in the time domain (Fig.: \ref{fig:LSTMtestSmall}). To evaluate the similarity with the smoothed original muscle activity and the difference between all artificially generated signals, the MSE \eqref{eq:MSE} is computed, which was also used as a loss function by the neural networks. Due to the over-representation of lower values caused by the rest state and inactive muscle groups for most motions, the MSE is inherently lower values than naturally expected. Note that the MSE of the original data is only $0.001$ while the signal is allowed to have values up to $1$. To account for this, we introduce a new rating, the zero-line score (Z-score) \eqref{eq:Z-score}, calculating a zero line signal comparison. With the mean square value of the original signal MSE\textsubscript{0} \eqref{eq:MSE_0} as a baseline, a score of $0$ indicates an approximation as poor as the zero line signal itself, while a value of 100 signifies perfect alignment. The Z-score is especially useful for capturing the error of muscle activity for multiple channels, which are of different magnitude.
\begin{align}
    \phantom{xxxxxxxxxxxxxxxxxxxxxxxxxxxxx}\text{MSE} &= \frac{1}{n}\sum_{i=1}^{n} (y_i-x_i)^2,\label{eq:MSE}\\[1ex]
    \text{MSE\textsubscript{0}} &=\frac{1}{n}\sum_{i=1}^{n}(y_i-0)^2,\label{eq:MSE_0}\\[1ex]
    \text{Z\textsubscript{s}} &= 100 \cdot\biggl(1-\frac{\text{MSE}}{\text{MSE\textsubscript{0}}}\biggr)\,.\label{eq:Z-score}
\end{align}

Note that the Z-score can easily reach values smaller than zero e.g. when the predicted signal is higher than twice the original signal. The following Section presents the results for the different architectures and training inputs.

\subsection{Comparison of the network architectures}\label{sec:Architecture}
Across all architectures, the artificial muscle activity is approximated reasonably close. In the following, the architectures are evaluated based on a general model containing train and test data from multiple subjects. Thereby the recurrent neural network outperforms the others with a score of $88.13$ (Tab.: \ref{tab:architecture}). The sub-sequence based online recurrent neural network achieves the lowest score of $83.33$. The scores for each channel reveal a higher accuracy for the second and third channels, which represent the deltoid muscle activity. 

\begin{table}[H]
\caption{Overview of the performance for all architectures (recurrent neural network (RNN), sub-sequenced input recurrent neural network (RNNseq), feedforward neural network (FNN), sub-sequenced input feedforward neural network (FNNseq), convolutional neural network (CNN)) and channels (electrodes 1-8) with the general approach (training and testing on multiple subjects) evaluated by the zero-line score (Z\textsubscript{s}) and mean square error (MSE).} \vspace{0.5cm}
\begin{tabular*}{\textwidth}{ @{\extracolsep{\fill}} l|l|llllllll|l}
                         &  & 1  & 2  & 3  & 4  & 5  & 6  & 7  & 8  & average \\ 
                        \midrule 
\multirow{2}{*}{RNN}    & Z\textsubscript{s} & 89.53  & 91.89   & 90.31   & 88.75   & 81.27   & 89.45   & 80.55   & 80.44   & \textbf{88.13}   \\
                        & MSE & 0.0009 & 0.00131 & 0.00285 & 0.00203 & 0.00269 & 0.00066 & 0.00066 & 0.00081 & 0.00149 \\
                        \midrule 
\multirow{2}{*}{RNNseq} & Z\textsubscript{s} & 86.24   & 90.34   & 86.78   & 84.71   & 69.16   & 84.42   & 76.11   & 72.94   & 83.33   \\ 
                        & MSE & 0.00119 & 0.00156 & 0.00388 & 0.00275 & 0.00443 & 0.00098 & 0.00081 & 0.00113 & 0.00209 \\
                        \midrule 
\multirow{2}{*}{FNN}    & Z\textsubscript{s} & 88.39 & 90.97   & 90.06   & 88.43   & 63.58   & 88.59   & 79.81   & 82.19   & 85.21   \\ 
                        & MSE & 0.001 & 0.00146 & 0.00292 & 0.00208 & 0.00523 & 0.00071 & 0.00068 & 0.00074 & 0.00185 \\
                        \midrule 
\multirow{2}{*}{FNNseq} & Z\textsubscript{s} & 88.6    & 91.67   & 89.74   & 88.57   & 75.35   & 86.73   & 78.21   & 81.71   & 86.81  \\   
                        & MSE  & 0.00102 & 0.00139 & 0.00311 & 0.00214 & 0.00356 & 0.00085 & 0.00077 & 0.00078 & 0.0017  \\
                        \midrule 
\multirow{2}{*}{CNN}    & Z\textsubscript{s} & 88.49   & 91.24   & 89.78 & 88.25   & 71.94   & 86.59   & 79.89   & 81.0    & 86.18   \\
                        & MSE & 0.00099 & 0.00141 & 0.003 & 0.00212 & 0.00403 & 0.00084 & 0.00068 & 0.00079 & 0.00173
\label{tab:architecture}
\end{tabular*}
\end{table}

\subsection{Performance for new subjects}\label{sec:Individual training inputs}
The next step is to evaluate how the model predicts new (unseen) subjects. For this purpose, the general model from the Section \ref{sec:Architecture} is first tested on a separate subject. In this setup, all architectures reach a significantly lower similarity between $30$ and $40$ (Tab.:\ref{tab:VS}). This is not surprising due to inter-subject variance for further information see Section \ref{sec:Discussion}. 

To improve the performance, we apply a subject-specific fine-tuning on the general model using transfer learning with a weight initialization strategy \citep{pan2009survey}. Thereby, the weights of the original general model are used as a pre-training status which is then updated on the subject-specific data. This approach leads to a significant increase in performance for all architectures and reaches similar values to the multiple-subject setting described in the Section \ref{sec:Architecture} with a slightly higher maximum of $88.21$ for the RNN (Tab.: \ref{tab:VS}). For comparison, we further train a model purely on one subject's data (data from the previously separate subject), resulting in a subject-specific model. This model again has a high performance of over $80$ across most architectures (Tab.: \ref{tab:VS}) but remains behind the pre-trained model.     

\begin{table}[H]
\caption{All architectures (recurrent neural network (RNN), sub-sequenced input recurrent neural network (RNNseq), feedforward neural network (FNN), sub-sequenced input feedforward neural network (FNNseq), convolutional neural network (CNN)) are tested on the same test data set of the separate subject and evaluated by the zero-line score (Z\textsubscript{s}) and mean square error (MSE).} \vspace{0.5cm}
\centering
\begin{tabular}{l|l|lll}  
                        &         & general & pre-train   & subject-specific  \\ 
                        \midrule 
\multirow{2}{*}{RNN}    & Z\textsubscript{s}       & 35.52   & \textbf{88.21}   & 85.16      \\
                        & MSE       & 0.00747 & 0.00137 & 0.00172  \\
                        \midrule 
\multirow{2}{*}{RNNseq} & Z\textsubscript{s}       & 41.23   & 83.36   & 77.12      \\
                        & MSE       & 0.00681 & 0.00193 & 0.00265   \\
                        \midrule 
\multirow{2}{*}{FNN}    & Z\textsubscript{s}       & 36.68   &  83.83    &  85.74     \\
                        & MSE       & 0.00733 & 0.00187   & 0.00165  \\
                        \midrule 
\multirow{2}{*}{FNNseq} & Z\textsubscript{s}     & 40.71  & 87.95   & 85.67      \\
                        & MSE       & 0.00714  & 0.00145 & 0.00173   \\
                       \midrule 
\multirow{2}{*}{CNN}    & Z\textsubscript{s}       & 42.61   & 80.18   & 86.56   \\
                        & MSE       & 0.00665 & 0.0023 & 0.00153     
\label{tab:VS}
\end{tabular}
\end{table}

\subsection{Generalization property for new motion}\label{sec:Generating new motion}
Thus far, we have shown that we can predict muscle activity with different architectures and also generate subject-specific muscle activities. The natural consequence of this is to go beyond and generate new motion that has not been learned before. For this, all models get tested on a new motion not seen before, pointing to 3 points in space (Fig.: \ref{fig:motion3Pmotion}) for the separated subject introduced in Section \ref{sec:Individual training inputs}. Overall, the pre-trained models reach a consistently higher accuracy than the general model, led by the RNN and CNN with $71.6$  (Tab.: \ref{tab:newMotion}). In all cases, the pre-trained models also outperform the subject-specific model. 

\begin{table}[H]
\caption{All architectures (recurrent neural network (RNN), sub-sequenced input recurrent neural network (RNNseq), feedforward neural network (FNN), sub-sequenced input feedforward neural network (FNNseq), convolutional neural network (CNN)) are tested on a new motion from the separate subjects and evaluated by the zero-line score (Z\textsubscript{s}) and mean square error (MSE).}  \vspace{0.5cm}
\centering
\begin{tabular}{l|l|lll}  
                        &         & general & pre-train   & subject-specific  \\ 
                        \midrule 
\multirow{2}{*}{RNN}    & Z\textsubscript{s}       & 53.42   & \textbf{71.6}   & 71.33      \\
                        & MSE       & 0.01286 & 0.00784 & 0.00791  \\
                        \midrule 
\multirow{2}{*}{RNNseq} & Z\textsubscript{s}       & 49.22   & 71.27   & 67.39      \\
                        & MSE       & 0.01401 & 0.00793 & 0.009   \\
                        \midrule 
\multirow{2}{*}{FNN}    & Z\textsubscript{s}       & 50.39   &  67.33    &  64.76     \\
                        & MSE       & 0.01369 & 0.00902   & 0.00973  \\
                        \midrule 
\multirow{2}{*}{FNNseq} & Z\textsubscript{s}     &37.51  & 70.99   & 69.47      \\
                        & MSE       &0.01786  & 0.00829 & 0.00873   \\
                       \midrule 
\multirow{2}{*}{CNN}    & Z\textsubscript{s}       & 47.87   & \textbf{71.63}   & 68.22   \\
                        & MSE       & 0.01439 & 0.00783 & 0.00877  
\label{tab:newMotion}
\end{tabular}
\end{table}

\begin{figure}[H]
\centering
\includegraphics[width=\textwidth]{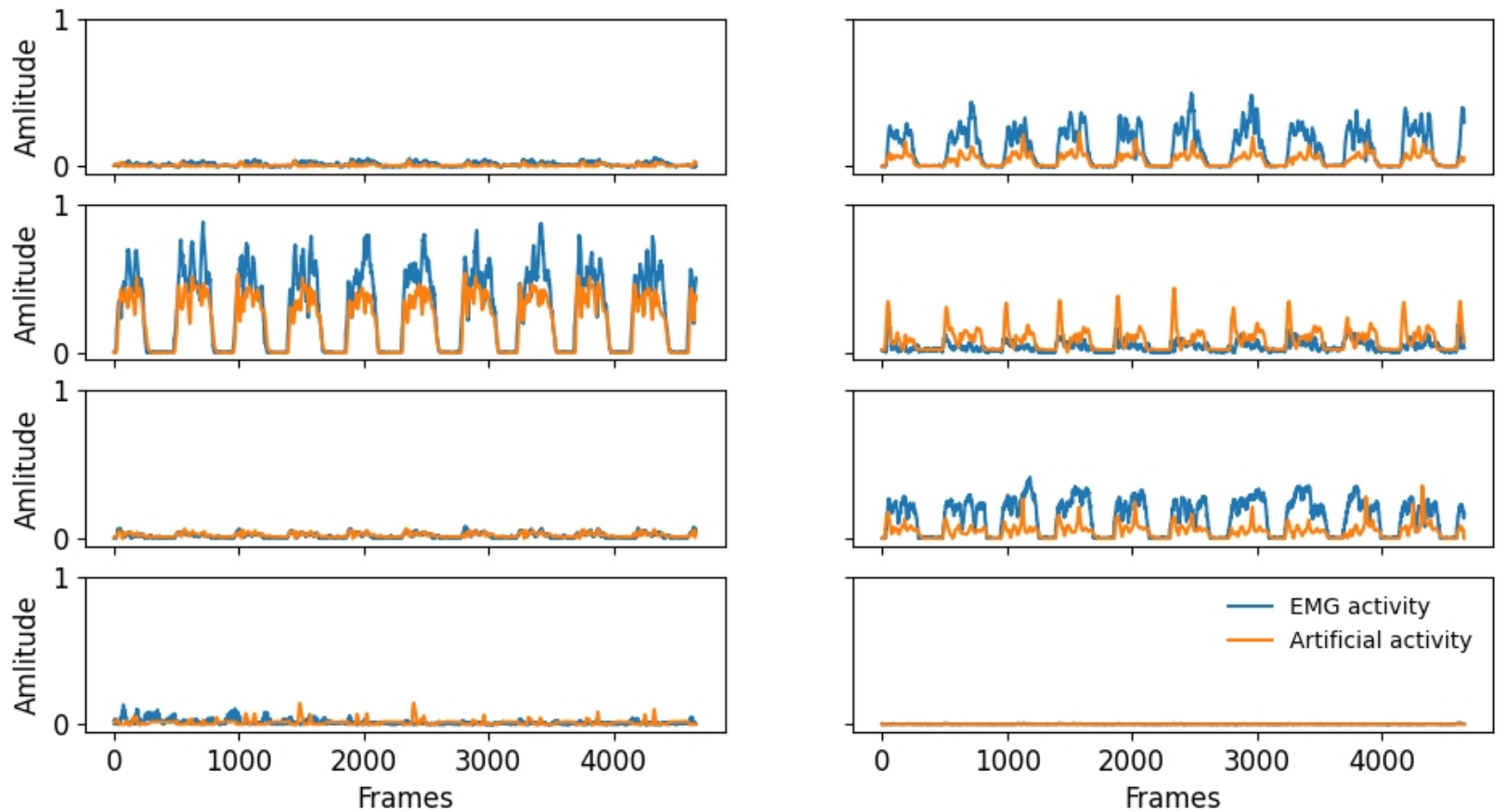}
\caption{Muscle activity from a new motion (pointing into 3 points in space) for all eighth channels (blue) and the artificial generated muscle activity (orange). The artificial activity is generated on the general training approach using a RNN architecture.}
\label{fig:LSTMtestSmallFree} 
\end{figure}

\subsection{Input parameter validation}\label{sec:Input validation}
Until now, we generated artificial muscle activity for known and new motions based on the chosen input parameter angular position, angular velocity, and angular acceleration for each joint (shoulder, elbow, and wrist). However, from an analytical viewpoint, the latter parameters are redundant since velocity and acceleration can be derived as the first and second derivatives of the position, respectively. Note that with integration on one, the velocity and position can be accumulated from the acceleration, and thus up to numerical precision, only one of the three quantities is needed to describe motion. Therefore, we want to test whether one of these three parameters is sufficient enough as input for a neural network to provide a similar good abstraction for generating known (Tab.: \ref{tab:VSangvelacceef}) and new muscle activity (Tab.: \ref{tab:freeangvelacceef}). To this end, we train the RNN separately on each input parameter and compare the ability to generate motions to the previous results using all three input quantities simultaneously. Furthermore, instead of testing the motion for each joint as before, we want to consider the position and orientation of the hand as input data, the so-called end effector.

The RNN trained with all input parameters still outperforms all other approaches with a reduced number of input data for a new subject and new motion (Tab.: \ref{tab:freeangvelacceef}, \ref{tab:VSangvelacceef}). The result is particularly evident in the case of the new motion prediction (Tab.: \ref{tab:freeangvelacceef}). For the general and pre-train approaches, the accuracy drops from a score of $71.55$ for solely using the angular position to $57.49$ for the acceleration input. A similar trend can be observed for the subject-specific models. The EEF and EEF\textsuperscript{+} achieve comparable results to the RNN trained with all parameters. However, the subject-specific model slightly outperforms the other models with a score of $71.98$.   

\begin{table}[H]
\caption{The recurrent neural network (RNN) is separately trained on several parameters: angular position (ang), angular velocity (vel) and angular acceleration (acc) from each joint. As well as on the end-effector (EEF) position and orientation of the hand and EEF\textsuperscript{+} with additional velocity and acceleration. The RNN is tested on a new motion from the separate subjects and evaluated by zero-line score (Z\textsubscript{s}) and mean square error (MSE).}

\centering
\begin{tabular}{l|l|lll}  
RNN                     &           & general   & pre-train & subject-specific  \\ 
                        \midrule 
\multirow{2}{*}{all}    & Z\textsubscript{s}       & 53.42   & \textbf{71.6}   & 71.33      \\
                        & MSE       & 0.01286 & 0.00784 & 0.00791  \\
                        \midrule  
\multirow{2}{*}{ang}    & Z\textsubscript{s}       & 36.49     & 71.55      & 61.38   \\
                        & MSE       & 0.01753   & 0.00785   & 0.01066 \\
                        \midrule 
\multirow{2}{*}{vel}    & Z\textsubscript{s}       & 45.22     & 60.7     & 60.4   \\
                        & MSE       & 0.01512   & 0.01085   & 0.01093  \\
                        \midrule 
\multirow{2}{*}{acc}    & Z\textsubscript{s}       & 40.3     & 57.49     & 58.96 \\
                        & MSE       & 0.01648   & 0.01173   &  0.01133\\
                        \midrule
\multirow{2}{*}{EEF}    & Z\textsubscript{s}     &  59.26   & 69.4   & \textbf{71.98}  \\
                        & MSE       & 0.01124 & 0.00845 &  0.00773 \\
                        \midrule
\multirow{2}{*}{EEF\textsuperscript{+}} & Z\textsubscript{s}       & 42.71 & 71.44   & 70.79   \\
                        & MSE       & 0.01581 & 0.00788 & 0.00806
                        
\label{tab:freeangvelacceef}
\end{tabular}
\end{table}

\section{Discussion}\label{sec:Discussion}  
The underlying motivation of this work is to demonstrate that artificial muscle activity of known and unknown motion can be generated based on motion parameters such as angular position, acceleration, and velocity of each joint (or the end-effector instead), which are similarly represented in our brain. For this purpose, we develop a neural network with a recurrent architecture that is trained in a supervised learning session. Alternative architectures are also elaborated and tested for comparison. Furthermore, we evaluate different training approaches: the general model, the pre-trained one, and the subject-specific model. All architectures and the majority of the other training approaches produce good artificial muscle activity for previously trained movements. In addition, we also generate artificial muscle activity on new motion, i.e., types of motions that were previously not used to train the network. Naturally, this is a much more challenging task, and trained motions achieve higher similarities than these new motions. 

The general setting of comparing different neural network architectures is described in Section \ref{sec:Architecture}. The RNN outperforms the other architectures for most muscle activity. In all models, the Z-score is higher for the second and third channels due to an unbalanced amount of data. These channels represent the shoulder abduction and flexion, which have a higher presentation in the recorded motion, whereas wrist motions are less represented and tend to have lower approximation values (channels seven and eight).

The general model does not generalize well for new subjects data (Tab.:\ref{tab:VS}). That is most likely due to the high inter-subject variation of the recorded muscle activity. To minimize these differences primarily caused by varying skin conditions a normalization factor is calculated and applied for each subject \citep{halaki2012normalization}. However, the normalization only accounts for a linear relationship between the subjects; in addition, there are likely to be additional changes in shape caused by individual anatomy and slightly inconsistent placement of the sensors \citep{sheng2019common, farina2014extraction, araujo2000inter, nordander2003influence, hogrel1998variability, farina2002influence}. Thus we must further enhance the model so that it can more accurately predict new subjects' data as well. The new model (pre-trained model) starts with the weights of the general model, which are further fine-tuned by an additional short training session with the data of the new subject utilizing transfer learning \citep{pan2009survey}. There are other pre-training approaches, however, weight initialization seems to be the most promising for muscle data \cite{lehmler_deep_2021}. Alternatively to the pre-trained model, a subject-specific model (purely trained on one subject's data) can also be used directly. The latter has fewer data for training compared to the pre-trained model strategy. Overall, the pre-trained model outperforms the subject-specific model (Tab.: \ref{tab:VS}) both for predicting artificially generated muscle activities of the new subject's data and, naturally, for previously used subjects.

The input parameter validation reveals that a combination of angular position, velocity, and acceleration results in a slightly higher overall accuracy than each parameter on its own, especially by being more robust for new motions (Tab.: \ref{tab:freeangvelacceef}). This is consistent with findings in motor cortical activity during reaching where besides the movement direction also less dominant correlations velocity and acceleration are represented \citep{ashe1994movement, moran1999motor}. From an analytical point of view, the redundancy of input parameters, for example, velocity, should not increase the accuracy of the overall model. In contrast to biological systems, and especially the neuronal system, a high degree of redundancy is present. The artificial neural networks inspired by these also exhibit an increase in performance, stability, and faster convergence due to the application of redundancy \citep{medler1994using,izui1990analysis}. The EEF\textsuperscript{+} as an input parameter reaches similarly high scores for the pre-trained model as the RNN with angular position, velocity, and acceleration of each joint. Note that the variation in the EEF may be lower than the variation of the overall sum of all joints in the arm. That could be explained by the fact that the arm has seven degrees of freedom, whereas only six degrees are sufficient to describe the EEF. The former allows choosing slightly different trajectories ending at the same position and orientation of the hand. The addition of velocity and acceleration parameters as further input, which was previously shown beneficial, for the angular position of all joints, causes the EEF\textsuperscript{+} to be able to achieve slightly higher scores for the pre-trained model.

The next step is to see if the models can generate muscle activity based on new motions. This reveals the true ability to abstract the relationship between the motion parameters and muscle activity. As expected, the accuracy for the new motion is lower than for the trained motions. However, the artificial signal still follows the trend of the original signal well (Fig.: \ref{fig:LSTMtestSmallFree}).

Most models generate muscle activity using an entire motion sequence and thus work offline. The RNNseq and FNNseq model are based on sub-sequences and are thus suitable for online prediction, however, it leads to a slightly decrease in overall accuracy.  

The recorded muscle activity includes an over-representation of the zero line representing an inactive state of the corresponding muscle. Initially, the motion sequences are cut to eliminate inter-trial pauses. However, due to the different lengths of each single task, the sequence still incorporates some zero line signals to ensure that all sequences have the same length. Further, not all recorded muscles are active in each motion. Most tasks are designed to activate only certain muscle groups such that all the other channels have a resting signal close to zero. Ensuring the same signal length is especially important for the CNN, while the RNN and FNN can cope with different sequence lengths. 

While the representation of the zero line itself is meaningful, as it represents the non-active state which is crucial for learning, this over-representation of low values leads to inherently small values in the MSE metric, which can easily be misinterpreted. As a result, the zero lines are easily predicted by all models, but learning other values becomes more challenging, and the model naturally tends to form lower peak muscle activity values overall. 

The newly introduced zero-line score is a rating that accounts for the over-representation of the zero line by adjusting the resolution accordingly, i.e., 100 always indicates a perfect fit and 0 an approximation as poor as the zero line, respectively. The smaller the values of the original signal (causing a closer resolution of the Z-score) are, the more difficult it becomes to reach a high Z-score for the artificially generated signal. This is due to the fact that the smallest possible Z-score value decreases. For the data used to evaluate the general model, the Z-score can hypothetically attain a value of $-5000$. (It describes the error between the original signal and a hypothetical signal containing the respectively more distant limit 0 or 1). 

The RNN receives the highest score for all motions used for training with a Z-score of $88$, while new motions reach a value up to $72$. We emphasize that a theoretical $100$ cannot be achieved due to the generalization by training the networks with different subjects, as mentioned before. Furthermore, we already described above that it is more difficult to achieve a high Z-score for data containing many zero signal segments. Note that the MSE achieves very low values of $0.00066$ and $0.00131$ (Tab.: \ref{tab:architecture}), respectively. Last, even for a hypothetical perfect test environment in which a subject can perform two exactly identical movements, it is not clear whether the corresponding muscle activities must also be perfectly matched. Consequently, our models achieve a very high degree of similarity, as can also be seen in Figure \ref{fig:LSTMtestSmall}. 

\section{Outlook}\label{sec:Outlook}  
The aim of the project is to develop a generative model which can generate muscle activity based on motion data. For this purpose, we have specially recorded several isotonic motions and their corresponding muscle activities. In contrast, muscle activity that does not cause a motion such as isometric contraction is more challenging to generate due to missing velocity and acceleration measures. Thus, the prediction solely relies on positional information and time dependencies.

This work may further be valuable to support EMG-based classifiers for myoelectrical controlled devices with additional data and thus to increase performance and generalization. Furthermore, transfer to a functional electric stimulation (FES) protocol to support the movement of paralyzed limbs would also be conceivable. Here, the relation between muscle activity and force is approximated and then transferred to stimulation patterns. On the other hand, an inversely controlled arm model would be conceivable to produce desired motion features and directly generate the corresponding muscle activity to increase the number of trained motions.

We further work on this topic and improve the network approach to train on multiple networks for each individual muscle activity with the help of an additional classifier. Thereby, we can reduce the influence of the zero lines for non-required muscles for each movement and improve the overall result and the individual performance.

\section*{Acknowledgements}
This work is supported by the Ministry of Economics, Innovation, Digitization and Energy of the State of North Rhine-Westphalia and the European Union, grants GE-2-2-023A (REXO) and IT-2-2-023 (VAFES).

\section*{Author contributions statement}
M.D.S conceived and conducted all experiments as well as analyzed the results. The original manuscript was written by M.D.S and reviewed by all authors T.G., I.I..

\section*{Materials \& Correspondence}
Correspondence to Marie D. Schmidt

\section*{Competing interests statement}
The authors declare that they have neither a financial nor a non-financial competing interest.

\bibliography{Literatur,zoteroGroup}
\section{Appendix}

\begin{table}[H]
\caption{The recurrent neural network (RNN) is separately trained on several parameters angular position (ang), angular velocity (vel) and angular acceleration (acc) from each joint.As well as on the end-effector (EEF) position and orientation of the hand and EEF\textsuperscript{+} with additional velocity and acceleration. The RNN is tested on the same test data set of the separate subject and evaluated by the zero-line score (Z\textsubscript{s}) and mean square error (MSE).} 
\centering
\begin{tabular}{l|l|lll}  
RNN                     &           & general   & pre-train & subject-specific  \\ 
                        \midrule 
\multirow{2}{*}{all}    & Z\textsubscript{s}       & 35.52   & 88.21   & 85.16      \\
                        & MSE       & 0.00747 & 0.00137 & 0.00172  \\
                        \midrule  
\multirow{2}{*}{ang}    & Z\textsubscript{s}       & 28.93     & 80.16      & 72.86   \\
                        & MSE       & 0.01923   & 0.00823   & 0.0023 \\
                        \midrule 
\multirow{2}{*}{vel}    & Z\textsubscript{s}       & 27.0     & 84.72     & 69.24   \\
                        & MSE       & 0.00846   & 0.00177   & 0.00356  \\
                        \midrule 
\multirow{2}{*}{acc}    & Z\textsubscript{s}       & 5.66     & 84.08     & 77.18 \\
                        & MSE       & 0.01093   & 0.00184   & 0.00264 \\
                        \midrule
\multirow{2}{*}{EEF}    & Z\textsubscript{s}     &  39.81   & 87.57   & 83.56  \\
                        & MSE       & 0.00697 & 0.00144 &  0.0019 \\
                        \midrule
\multirow{2}{*}{EEF\textsuperscript{+}} & Z\textsubscript{s}       & 30.41 & \textbf{89.48}   & 86.37   \\
                        & MSE       & 0.00806 & 0.00122 & 0.00158
\label{tab:VSangvelacceef}
\end{tabular}
\end{table}
\end{document}